# Farthest-Point Heuristic based Initialization Methods for K-Modes Clustering

Zengyou He

Department of Computer Science and Engineering, Harbin Institute of Technology,

92 West Dazhi Street, P.O Box 315, Harbin 150001, P. R. China

zengyouhe@yahoo.com

**Abstract**
The *k*-modes algorithm has become a popular technique in solving categorical data clustering problems in different application domains. However, the algorithm requires random selection of initial points for the clusters. Different initial points often lead to considerable distinct clustering results. In this paper we present an experimental study on applying a farthest-point heuristic based initialization method to *k*-modes clustering to improve its performance. Experiments show that new initialization method leads to better clustering accuracy than random selection initialization method for *k*-modes clustering.

## 1. Introduction

The *k*-modes algorithm [1] extends the *k*-means paradigm to cluster categorical data by using (1) a simple matching dissimilarity measure for categorical objects, (2) modes instead of means for clusters, and (3) a frequency-based method to update modes in the *k*-means fashion to minimize the cost function of clustering. Because the *k*-modes algorithm uses the same clustering process as *k*-means, it preserves the efficiency of the *k*-means algorithm.

Although the *k*-modes algorithm is very efficient, it suffers the problem that the clustering results are sensitive to the selection of the initial points. Hence, a better initial points selection procedure would improve the reliability and accuracy of clustering results. To that end, an iterative initial-points refinement algorithm for *k*-modes clustering has been presented in [2].

As shown in [2], the new initialization procedure greatly improves the reliability and accuracy of final clustering results. Despite the success of Ref. [2], the following observations motivate us to further pursue other alternative initialization methods.

(1). The clustering output of iterative initial-points refinement algorithm studied in [2] is still "random" in nature. That is, different runs of the algorithm give different clustering results. To get confident clustering output, the end-user still has to execute the algorithm repetitively. As shown in the experimental results in [2], very poor clustering results can occur in some cases. Hence, non-randomized initialization algorithm is much desirable in real applications.

(2). The new initialization method should be simpler and easier to implement. As a result, it is expected that such initialization method deserves good scalability.

(3). It would be more desirable if the new initialization algorithm can provide performance guarantee to certain degree.

Motivated by the above observations, this paper presents two farthest-point heuristic based initialization methods, which are borrowed from *k*-center problem [8]. These two new initialization algorithms have the following salient features: (1) Only one initial point is selected randomly in the first algorithm and no points are selected randomly in the second algorithm. It indicates that fixed clustering results could be expected from our initialization procedure based *k*-modes clustering algorithm; (2) New initialization algorithms are simple and fast, which requires at most $O(nk)$ time, where *n* is number of objects in the dataset and *k* is number of desired clusters; (3) The initialization algorithms applied in this paper are originally used for solving *k*-center clustering, which provide an approximation factor of 2 for *k*-center clustering. Although the objective function of *k*-center clustering is different from that of *k*-modes clustering, the quality of initial points can be expected to be "good" from certain perspective.

As shown in the experimental study, the clustering performance of *k*-modes algorithm using farthest-point heuristic based initialization methods are better than that of most variants of *k*-modes algorithm.

## 2. Related Work

Our focus in this paper is to study *k*-modes type clustering for categorical data. Hence, we will review only those *k*-modes related papers. Following the *k*-modes algorithm, many research efforts [3-7] have been conducted to further improve its performance.

Huang and Ng introduced the fuzzy *k*-modes algorithm [3], which assigns membership degrees to data objects in different clusters. The tabu search technique is applied in [4] and genetic algorithm is utilized in [5] to improve *k*-modes algorithm. Alternatively, fuzzy *k*-modes algorithm is extended by representing the clusters of categorical data with fuzzy centroids instead of the hard-type centroids used in the original algorithm [6-7]. However, most of these methods are much slower than the original *k*-modes algorithm in running time.

Since the *k*-modes algorithm is sensitive to the initial conditions, another feasible way for improving its performance is to design effective initialization methods. To that end, an iterative initial-points refinement algorithm for *k*-modes clustering is presented in [2].

## 3. The *k*-modes Algorithm

Let $D = \{X_1, X_2, \ldots, X_n\}$ be a set of *n* categorical objects/points, where each $X_i = [x_{i,1}, x_{i,2}, \ldots, x_{i,m}]$ is described by *m* categorical attributes $A_1, \ldots, A_m$. Each attribute $A_j$ describes a domain of values, denoted by $Dom(A_j) = \{a_j^{(1)}, a_j^{(2)}, \ldots, a_j^{(p_j)}\}$, where $p_j$ is the number of category values of attribute $A_j$.

The *k*-modes algorithm uses the *k*-means paradigm to search a partition of *D* into *k* clusters that minimize the objective function *P* with unknown variables *U* and *Z* as follows:

$$P(U,Z) = \sum_{l=1}^{k}\sum_{i=1}^{n}\sum_{j=1}^{m} u_{i,l} d(x_{i,j}, z_{l,j}) \qquad (1)$$

Subject to $\sum_{l=1}^{k} u_{i,l} = 1, \quad 1 \le i \le n$ (2)

$$u_{i,l} \in \{0,1\}, \quad 1 \le i \le n, \quad 1 \le l \le k \qquad (3)$$

where

- $U$ is an $n \times k$ partition matrix, $u_{i,l}$ is a binary variable, and $u_{i,l} = 1$ indicates that object $X_i$ is allocated to cluster $C_l$;
- $\mathbf{Z} = \{Z_1, Z_2, \ldots, Z_k\}$ is a set of vectors representing the centers of the $k$ clusters, where $Z_l = [z_{l,1}, z_{l,2}, \ldots, z_{l,m}]$ ($1 \le l \le k$);
- $d(x_{i,j}, z_{l,j})$ is a distance or dissimilarity measure between object $X_i$ and the center of cluster $C_l$ on attribute $A_j$. In $k$-modes algorithm, a simple matching distance measure is used. That is, the distance between two distinct categorical values is 1, while the distance between two identical categorical values is 0. More precisely,

$$d(x_{i,j}, z_{l,j}) = \begin{cases} 0 & (x_{i,j} = z_{l,j}) \\ 1 & (x_{i,j} \ne z_{l,j}) \end{cases} \qquad (4)$$

The optimization problem in $k$-modes clustering can be solved by iteratively solving the following two minimization problems:

1. Problem $P_1$: Fix $Z = \widehat{Z}$, solve the reduced problem $P(U, \widehat{Z})$,

2. Problem $P_2$: Fix $U = \widehat{U}$, solve the reduced problem $P(\widehat{U}, Z)$.

Problem $P_1$ and $P_2$ are solved according to the two following theorems, respectively.

**Theorem 1**: Let $Z = \widehat{Z}$ be fixed, $P(U, \widehat{Z})$ is minimized iff

$$u_{i,l} = \begin{cases} 1 & (\sum_{j=1}^{m} d(x_{i,j}, z_{l,j}) \le \sum_{j=1}^{m} d(x_{i,j}, z_{h,j}), \forall h, 1 \le h \le k) \\ 0 & (otherwise) \end{cases}$$

**Theorem 2**: Let $U = \widehat{U}$ be fixed, $P(\widehat{U}, Z)$ is minimized iff

$$z_{l,j} = a_j^{(r)}$$

where $a_j^{(r)}$ is the mode of attribute values of $A_j$ in cluster $C_l$ that satisfies $f(a_j^{(r)} | C_l) \ge f(a_j^{(t)} | C_l), \forall t, 1 \le t \le p_j$. Here $f(a_j^{(r)} | C_l)$ denotes the frequency count of attribute value $a_j^{(r)}$ in cluster $C_l$, i.e., $f(a_j^{(r)} | C_l) = |\{u_{i,l} | x_{i,j} = a_j^{(r)}, u_{i,l} = 1\}|$.

## 4. Farthest-Point Heuristic Based Initialization Methods

The *k*-center clustering problem is also called minmax radius clustering problem, whose objective is to minimize the maximum diameter of any cluster on some set of points. A simple 2-approximation algorithm for the *k*-center clustering problem is proposed by Gonzales [8], which utilizes a farthest-point clustering heuristic. In this paper, the algorithm and its variant are used as the initialization methods for *k*-modes clustering.

The farthest-point heuristic starts with an arbitrary point $s_1$. Pick a point $s_2$ that is as far from $s_1$ as possible. Pick $s_i$ to maximize the distance to the nearest of all centroids picked so far. That is, maximize the *min* {*dist* ($s_i$, $s_1$), *dist* ($s_i$, $s_2$), ...}. After all *k* representatives are chosen we can define the partition of *D*: cluster $C_j$ consists of all points closer to $s_j$ than to any other representative.

As shown in [8], let the maximum radius of these *k* clusters be $\sigma$, then even in the optimal *k*-clustering the maximum radius must be at least $\sigma/2$. Here optimal is in the sense of minimizing the maximum radius of any cluster.

Obviously, farthest-point heuristic based method has the time complexity *O* (*nk*), where *n* is number of objects in the dataset and *k* is number of desired clusters. Furthermore, one can implement farthest-point heuristic based method in constant dimension in *O* (*n*log*k*) time [9]. In fact, it can be solved in *O* (*n*) this case [10]. Clearly, farthest-point heuristic based method is fast and suitable for large-scale data mining applications.

In the above basic farthest-point heuristic (BFPH), the first point is selected randomly. Then, the remaining *k*-1 points are selected deterministically. Although only one point is selected randomly, it is also hard to guarantee stable clustering results if above basic farthest-point heuristic is used in *k*-modes clustering. In the following, we present a method for selecting the first point deterministically.

For each $X_i = [x_{i,1}, x_{i,2}, \ldots, x_{i,m}]$ in *D* that is described by *m* categorical attributes, we use $f(x_{i,j} | D)$ to denote the frequency count of attribute value $x_{i,j}$ in the dataset. Then, a scoring function is designed for evaluating each point, which is defined as: $socre(X_i) = \sum_{j=1}^{m} f(x_{i,j} | D)$.

In the new farthest-point heuristic (NFPH), the point with highest score is selected as the first point, and remaining points are selected in the same manner as that of basic farthest-point heuristic [8]. The rationale behind the scoring function is that, points that contain attribute values with higher frequencies are more likely to be centers. Obviously, in the new farthest-point heuristic, all points are selected deterministically. It indicates that fixed clustering results could be achieved from our new initialization procedure based *k*-modes clustering algorithm.

Moreover, selecting the first point according to above defined scoring function could be fulfilled in *O* (*n*) time by deploying the following procedure (with two scans over the dataset):

(1). In the first scan over the dataset, we construct *m* hash tables as our basic data structures to store the information on attribute values and their frequencies.

(2). In the second scan over the dataset, with the use of hashing technique, in *O* (1) expected time, we can determine the frequency count of an attribute value in corresponding hash table. Therefore, the data point with largest score could be detected in *O* (*n*) time.

Hence, the time complexity of new farthest-point heuristic (NFPH) is still $O(nk)$.

## 5. Experimental Results

We ran both BFPH and NFPH based *k*-modes algorithms on real-life datasets obtained from the UCI Machine Learning Repository [11] to test their clustering performance against original *k*-modes algorithm using random initialization.

### 5.1 Real Life Datasets and Evaluation Method

Four data sets from the UCI Repository are used, all of which contains only categorical attributes and class attributes. The information about the data sets is tabulated in Table 1. Note that the class attributes of the data have not been used in the clustering process.

**Table 1**. Datasets used in experiments

| Data set | Size | Attribute | Class | Class Distribution |
|---|---|---|---|---|
| Voting | 435 | 16 | 2 | 168/267 |
| Mushroom | 8124 | 22 | 2 | 3916/4208 |
| Soybean | 47 | 35 | 4 | 10/10/10/17 |
| Zoo | 101 | 17 | 7 | 4/5/8/10/13/20/41 |

Validating clustering results is a non-trivial task. In the presence of true labels, as in the case of the data sets we used, the clustering accuracy for measuring the clustering results was computed as follows [1]. Given the final number of clusters, *k*, clustering accuracy *r* was defined as: $r = \frac{\sum_{l=1}^{k} a_l}{n}$, where *n* is the number of objects in the dataset, $a_l$ is the number of instances occurring in both cluster $C_l$ and its corresponding class, which had the maximal value. In other words, $a_l$ is the number of objects with the class label that dominates cluster $C_l$.

The intuition behind clustering accuracy defined above is that clusterings with "pure" clusters, i.e., clusters in which all objects have the same class label, are preferable. That is, if a partition has clustering accuracy equal to 100%, it means that it contains only pure clusters. These kinds of clusters are also interesting from a practical perspective. Hence, we can conclude that larger clustering accuracy implies better clustering results in real world applications.

### 5.2 Results

For each dataset, the number of clusters is set to be the known number of its class labels. For instance, the number of clusters is set to be 2 on *voting* data. We carried out 100 random runs of the original *k*-modes and BFPH based *k*-modes algorithm on each data set. Since the NFPH based *k*-modes algorithm has deterministic and same output in every run, only one run is used for this algorithm. The average clustering accuracies of different algorithms were compared.

Table 2 lists the average accuracy of clustering achieved by each algorithm over 100 runs for the four data sets. From Table 2, some important observations are summarized as follows.

(1) Firstly, it is evident from Table 2 that all farthest-point heuristic based $k$-modes algorithms give better clustering accuracy in comparison to standard $k$-modes algorithm with random initialization. Hence, we can conclude that clustering accuracy could be greatly improved with the use of farthest-point heuristic as initialization technique in $k$-modes clustering.

(2) Secondly, clustering accuracy achieved by NFPH based $k$-modes algorithm is better than that of BFPH based $k$-modes algorithm in most cases. In particular, when the number of clusters $k$ is relatively smaller (e.g., the results on *voting* and *mushroom* data), NFPH is always better than BFPH in producing better clustering results. The reason is that the "randomness" of BFPH based initialization method will be increased if $k$ is decreased. More precisely, when $k$=2, one point is selected randomly and another point is selected deterministically. That is, 50% of points are selected randomly in BFPH in case of $k$=2. On the other hand, NFPH based $k$-modes algorithm and BFPH based $k$-modes algorithm have similar performance when $k$ is larger (e.g., the results on *soybean* and *zoo* data).

(3) Finally, *soybean* data and *zoo* data have frequently been used to test categorical clustering algorithms, we take those reported results directly from corresponding research papers for the purpose of comparison.

With respect to *soybean* data, the clustering accuracies achieved by other state-of-the-art $k$-modes type algorithms are 89.4% [2], 99.1% [4] and 100% [6] respectively. Clearly, our farthest-point heuristic is much better than iterative initial-points refinement algorithm [2] in initializing cluster centers in $k$-modes clustering. Furthermore, although the algorithms presented in [4] and [6] give competitive clustering accuracies as our methods, it should be noted that these methods are much slower in running time.

With respect to *zoo* data, the algorithm presented in [6] provides an accuracy of 86.58%, which is at least 6% less accurate than our methods.

**Table 2**. Average clustering accuracy (%) achieved by three algorithms on four datasets

| Data set | Standard $k$-modes | BFPH based $k$-modes | NFPH based $k$-modes |
|---|---|---|---|
| Voting | 85.92 | 85.27 | **86.44** |
| Mushroom | 73.81 | 77.64 | **80.00** |
| Soybean | 81.94 | 98.57 | **100.00** |
| Zoo | 82.92 | **93.02** | 92.08 |
| *Avg.* | *81.15* | *88.63* | ***89.63*** |

# 6. Conclusions

The conventional $k$-modes algorithm is efficient and effective in clustering large categorical data. However, its use of random initialization compromises its effectiveness and its ability to correctly classify categorical data. Therefore, this paper extends $k$-modes clustering algorithm by providing two farthest-point heuristic based initialization methods. The superiority of our methods was demonstrated through several experiments.

# References

1. Z. Huang. Extensions to the k-Means Algorithm for Clustering Large Data Sets with Categorical Values. Data Mining and Knowledge Discovery, 1998, 2: 283-304
2. Y. Sun, Q. Zhu, Z. Chen. An iterative initial-points refinement algorithm for categorical data clustering. Pattern Recognition Letters, 2002, 23(7): 875-884.
3. Z. Huang, M. K. Ng. A fuzzy k-modes algorithm for clustering categorical data. IEEE Transaction on Fuzzy Systems, 1999, 7(4): 446-452.
4. M. K. Ng, J. C. Wong. Clustering categorical data sets using tabu search techniques. Pattern Recognition, 2002, 35(12): 2783-2790.
5. G. Gan, Z. Yang, J. Wu. A Genetic *k*-Modes Algorithm for Clustering Categorical Data. In: Proc. of ADMA'05, pp.195-202, 2005
6. D-W. Kim, K. Y. Lee, D. Lee, K. H. Lee. A *k*-populations algorithm for clustering categorical data. Pattern Recognition, 2005, 38(7): 1131-1134.
7. D-W. Kim, K. H. Lee, D. Lee. Fuzzy clustering of categorical data using fuzzy centroids. Pattern Recognition Letters, 2004, 25(11): 1263-1271.
8. T.F. Gonzales. Clustering to minimize the maximum intercluster distance. Theoretical Computer Science, 1985, 38(2-3): 293-306.
9. T. Feder, D. H. Greene. Optimal algorithms for approximate clustering. In: Proc. of 20th Annual ACM Symposium on Theory of Computing (STOC'98), pp. 434-444, 1988.
10. S. Har-Peled. Clustering motion. Discrete & Computational Geometry, 2004, 31(4): 545–565.
11. Merz, C. J., Merphy. P.: UCI Repository of Machine Learning Databases. [http://www.ics.uci.edu/~mlearn/MLRRepository.html] (1996)